\PassOptionsToPackage{table}{xcolor}
\documentclass[10pt,twocolumn,letterpaper]{article}

\usepackage[pagenumbers]{iccv} 

\definecolor{iccvblue}{rgb}{0.21,0.49,0.74}
\usepackage[pagebackref,breaklinks,colorlinks,allcolors=iccvblue]{hyperref}

\usepackage{booktabs} 

\usepackage{pifont}

\definecolor{mygray}{gray}{.95}
\definecolor{highlight}{RGB}{238,250,215}
\definecolor{lightgray}{gray}{0.7}
\definecolor{lightorange}{HTML}{C47955}
\definecolor{darkgreen}{HTML}{195228}
\definecolor{eccvblue}{rgb}{0.12,0.49,0.85}
\definecolor{lightblue}{rgb}{0.12,0.49,0.85}

\def\paperID{*****} 
\def\confName{ICCV}
\def\confYear{2025}

\title{PRVQL: Progressive Knowledge-guided Refinement for Robust Egocentric \\ Visual Query Localization}

\author{
  Bing Fan$^{1}$ \quad
  Yunhe Feng$^{1}$ \quad
  Yapeng Tian$^{2}$ \quad
  James Chenhao Liang$^{3}$ \quad Yuewei Lin$^{4}$\\
  Yan Huang$^{1}$ \quad
  Heng Fan$^{1}$ \\
  $^{1}$University of North Texas \quad
  $^{2}$University of Texas at Dallas \\
  $^{3}$U.S. Naval Research Laboratory \quad
  $^{4}$Brookhaven National Laboratory
}

\begin{document}
\maketitle

\begin{abstract}
Egocentric visual query localization (\emph{EgoVQL}) focuses on localizing the target of interest in space and time from first-person videos, given a visual query. Despite recent progressive, existing methods often struggle to handle severe object appearance changes and cluttering background in the video due to lacking sufficient target cues, leading to degradation. Addressing this, we introduce \textbf{PRVQL}, a novel \textbf{P}rogressive knowledge-guided \textbf{R}efinement framework for Ego\textbf{VQL}. The core is to continuously exploit target-relevant knowledge directly from videos and utilize it as guidance to refine both query and video features for improving target localization. Our PRVQL contains multiple processing stages. The target knowledge from one stage, comprising appearance and spatial knowledge extracted via two specially designed knowledge learning modules, are utilized as guidance to refine the query and videos features for the next stage, which are used to generate more accurate knowledge for further feature refinement. With such a progressive process, target knowledge in PRVQL can be gradually improved, which, in turn, leads to better refined query and video features for localization in the final stage. Compared to previous methods, our PRVQL, besides the given object cues, enjoys additional crucial target information from a video as guidance to refine features, and hence enhances EgoVQL in complicated scenes. In our experiments on challenging Ego4D, PRVQL achieves state-of-the-art result and largely surpasses other methods, showing its efficacy. Our code, model and results will be released at \url{https://github.com/fb-reps/PRVQL}.
\end{abstract}

\section{Introduction}
\label{intro}

The egocentric visual query localization (EgoVQL) task~\cite{grauman2022ego4d} aims at answering the question ``\emph{Where was the object X last seen in the video?}'', with ``\emph{X}'' being a visual query specified by a single image crop outside the search video. In specific, given a first-person video, its goal is to search and locate the visual query, \emph{spatially} and \emph{temporally}, by returning the most recent spatio-temporal tube. Owing to its important roles in numerous downstream object-centric applications including augmented and virtual reality, robotics, human-machine interaction, and so on, EgoVQL has drawn extensive attention from researchers in recent years.

\begin{figure}[!t]
    \centering
    \includegraphics[width=\linewidth]{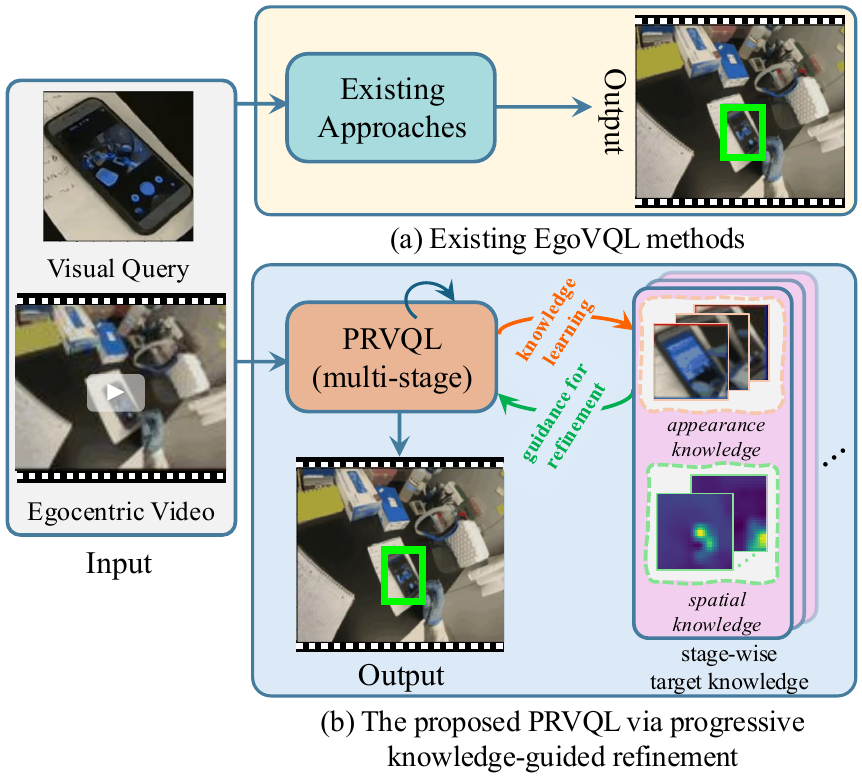}\vspace{-1mm}
    \caption{Comparison between current EgoVQL approaches in (a) and proposed PRVQL with progressive knowledge-guided refinement in (b). \emph{Best viewed in color and by zooming in for all figures}.}
    \label{fig:framework_comparison}\vspace{-2mm}
\end{figure}

Current approaches (\eg,~\cite{xu2022negative,xu2023my,jiang2024single,grauman2022ego4d}) simply leverage the provided visual query as the \emph{sole} cue to locate the target in the video (see Fig.~\ref{fig:framework_comparison} (a)). However, since the given visual query is cropped \emph{outside} the search video, there often exists a \emph{significant gap} between the query and the target of interest, due to rapid appearance variations in first-person videos caused by many factors, such as object pose change, motion blur, occlusion, and so forth. As a result, relying only on the given object query, as in existing methods, is \emph{insufficient} to describe and distinguish the target from background in complicated scenarios with heavy appearance changes, resulting in performance degeneration. In addition, to achieve precise localization, it is essential for an EgoVQL model to enhance target and meanwhile suppressing background regions from videos. Yet, this is often \emph{overlooked} by existing approaches, making them easily suffer from cluttering background and therefore leading to suboptimal target localization.

The aforementioned issues faced by current methods naturally raise a question: \emph{In addition to the given visual query, is there any other information that could be leveraged for enhancing EgoVQL}? We answer \emph{\textbf{yes}}, and show the information directly explored from the \emph{video itself}, as a supplement to the given target cue, is \emph{effective} in improving EgoVQL.

\begin{figure*}[!t]
	\centering
        \includegraphics[width=1\textwidth]{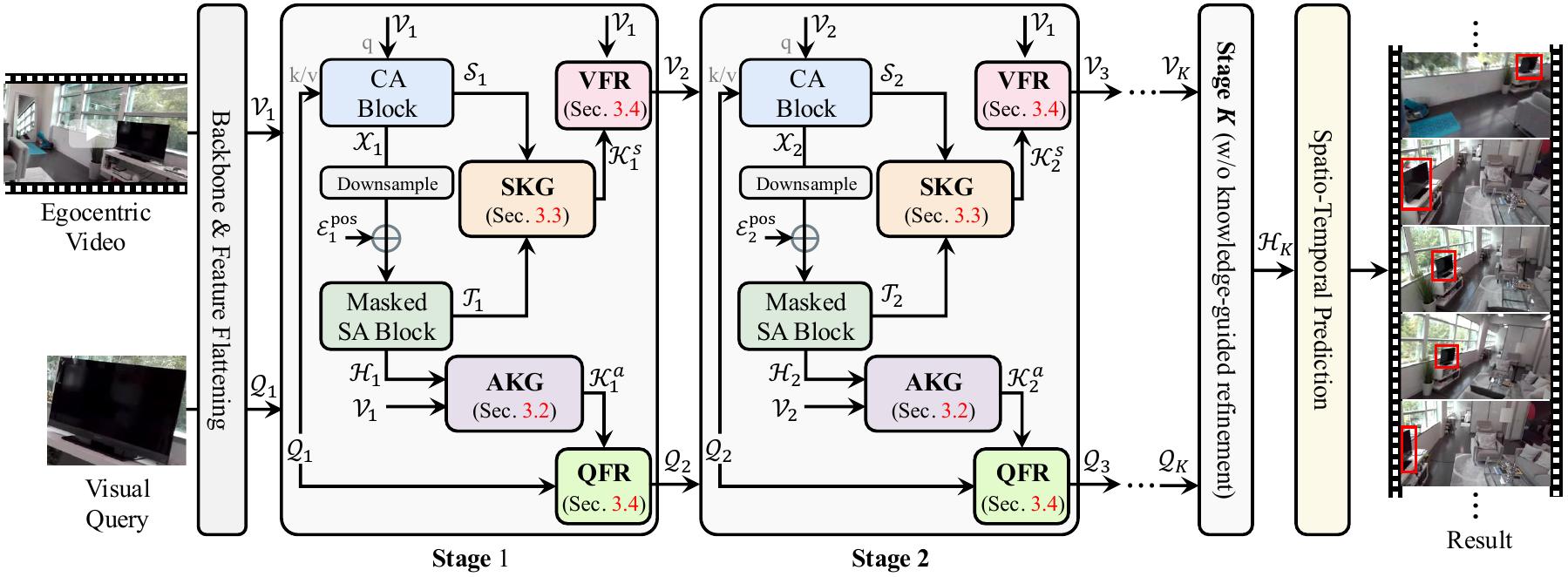}\vspace{-2mm}
	\caption{Overview of PRVQL, which aims to explore target knowledge directly from videos via AKG and SKG and applies it as guidance to refine query and video features with QFR and VFR for improving localization in EgoVQL through a multi-stage progressive architecture.}
	\label{fig:framework}\vspace{-4mm}	
\end{figure*}

Specifically, we propose a novel \emph{\textbf{P}rogressive knowledge-guided \textbf{R}efinement framework for Ego\textbf{VQL}} (\textbf{\emph{PRVQL}}). The core idea of our algorithm is to continuously exploit target-relevant knowledge from the video and leverage it to guide refinements of both query and video features, which are crucial for localization, for improving EgoVQL (see Fig.~\ref{fig:framework_comparison} (b)). Concretely, PRVQL consists of multiple processing stages. Each stage comprises two simple yet effective modules, including \emph{appearance knowledge generation} (AKG) and \emph{spatial knowledge generation} (SKG). In specific, AKG works to mine visual information of the target from videos as the appearance knowledge. It first estimates potential target regions from a video using the query, and then selects top few highly confident regions to extract appearance knowledge from video features. Different from AKG, SKG focuses on exploring target position cues from videos as spatial knowledge by exploiting the readily available target-aware attention maps. In PRVQL, the appearance knowledge is used to guide the update of query feature, making it more discriminative, while the spatial knowledge is employed to enhance target and meanwhile suppressing unconcerned background in video features, enabling more focus on the target. The extracted appearance and spatial knowledge in one stage are used as guidance to respectively refine query and video features for next stage, which are adopted to learn more accurate knowledge for further feature refinement. Through this progressive process in PRVQL, the target knowledge can be gradually improved, which, in turn, results in better refined query and video features for target object localization in the final stage. Fig.~\ref{fig:framework} illustrates the framework of PRVQL.

To our best knowledge, PRVQL is the \emph{first} method to exploit target-relevant appearance and spatial knowledge from the video to improve EgoVQL. Compared with existing solutions, PRVQL can leverage target information from both the given visual query and mined knowledge from the video for more robust localization. To verify its effectiveness, we conduct experiments on the challenging Ego4D~\cite{grauman2022ego4d}, and our proposed PRVQL achieves state-of-the-art performance and largely outperforms other approaches, evidencing effectiveness of target knowledge for enhancing EgoVQL. 

In summary, our main contributions are as follows: \ding{171} We propose a progressive knowledge-guided refinement framework, dubbed PRVQL, that exploits knowledge from videos for improving EgoVQL; \ding{170} We introduce AKG for exploring visual information of target as appearance knowledge; \ding{168} We introduce SKG for learning spatial knowledge using target-aware attention maps; \ding{169} In our extensive experiments on the challenging Ego4D, PRVQL achieves state-of-the-art performance and largely surpasses existing methods. 

\section{Related Work}


\textbf{Egocentric Visual Query Localization.} Egocentric visual query localization (EgoVQL) is an emerging and important computer vision task. Since its introduction in~\cite{grauman2022ego4d}, EgoVQL has received extensive attention in recent years owing to its importance in numerous applications. Early methods~\cite{grauman2022ego4d,xu2022negative,xu2023my} often utilize a bottom-up multi-stage framework, which sequentially and independently performs frame-level object detection, nearest peak temporal detection across the video, and bidirectional object tracking around the peak, to achieve EgoVQL. Despite the straightforwardness, this bottom-up design easily causes compounding errors across stages, thus degrading performance. Besides, the involvement of multiple detection and tracking components in this design leads to high complexities as well as inefficiency of the entire system, limiting its practicability. To deal with these issues, the recent method of~\cite{jiang2024single} introduces a single-stage end-to-end framework for EgoVQL with Transformer~\cite{VaswaniSPUJGKP17}, eliminating the need for multiple components and meanwhile showing promising target localization performance.

In this work, we propose to exploit target knowledge directly from the video and utilize it as guidance to refine features for better localization. \textbf{\emph{Different}} from aforementioned approaches~\cite{grauman2022ego4d,xu2022negative,xu2023my,jiang2024single} which mainly explore the object information from only the provided query for localization, PRVQL is able to leverage cues from both the given query and mined target information for EgoVQL, significantly improving robustness, especially in presence of severe appearance variations and cluttering background.

\vspace{0.5em}
\noindent
\textbf{Query-based Visual Localization.} Query-based visual localization, broadly referring to localizing the target of interest from images or videos given a specific query (image or text), is a crucial problem in computer 
vision, and consists of a wide range of related tasks, including one-shot object detection~\cite{hsieh2019one,yang2022balanced,zhao2022semantic}, visual object tracking~\cite{chen2023seqtrack,lin2025tracking,bertinetto2016fully}, visual grounding~\cite{deng2021transvg,liu2025grounding,zhu2022seqtr}, spatio-temporal video grounding~\cite{yang2022tubedetr,gu2024context}, pedestrian search~\cite{li2017person,yu2022cascade}, \etc. Despite sharing some similarity with the above tasks in localizing the target, this work is \textbf{\emph{distinctive}} by focusing on spatially and temporally searching for the target from egocentric videos, which is challenging due to frequent and heavy object appearance variations under the first-person views.

\vspace{0.5em}
\noindent
\textbf{Progressive Learning Approach.} Multi-stage progressive learning is a popular strategy to improve performance, and has been successfully applied for various tasks. For example, the works of~\cite{cai2018cascade,ye2023cascade,vu2019cascade} introduce the cascade architecture to progressively refine the bounding boxes or features for improving object detection. The work in~\cite{yang2019step} presents a sptio-temporal progressive network for video action detection. The methods in~\cite{huynh2021progressive,zhao2018icnet} introduce progressive refinement network for multi-scale semantic segmentation. The methods of~\cite{zhang2018progressive,chen2020progressively} apply progressive learning to improve features for saliency detection. The method in~\cite{fan2019siamese} proposes to progressively learn more accurate anchors for enhancing tracking. The work from~\cite{zhu2019progressive} progressively transfers person pose for image generation. \textbf{\emph{Different}} than these works, we focus on progressive refinement for improving EgoVQL.

\section{The Proposed Method}

\textbf{Overview.} In this paper, we propose PRVQL by exploiting crucial target knowledge directly from videos for improving target localization in EgoVQL. Our PRVQL is implemented as a progressive architecture. After feature extraction of the visual query and video frames, PRVQL performs iterative feature refinement guided by the target knowledge for localization through multiple stages (Sec.~\ref{prvql}). As displayed in Fig.~\ref{fig:framework}, each stage, expect for the final stage for prediction, consists of two crucial modules, comprising AKG (Sec.~\ref{akg}) and SKG (Sec.~\ref{skg}), for generating target appearance and spatial knowledge. The knowledge is leveraged as the guidance to refine query and video features (Sec.~\ref{update}), which are applied in the next stage to generate more accurate target knowledge for further feature refinement. Through such a progressive process, the target knowledge can be gradually enhanced, which finally benefits learning more discriminative query and video features for improving EgoVQL.

\subsection{Our PRVQL Framework}\label{prvql}

\textbf{Visual Feature Extraction.} In our PRVQL, we first extract features for the visual query and video frames. Specifically, given the query $q$ and a sequence of $L$ frames $\mathcal{I}=\{I_i\}_{i=1}^{L}$ from the video, we utilize a shared backbone $\Phi(\cdot)$~\cite{OquabDMVSKFHMEA24} for extracting their features $\textbf{q}=\Phi(q) \in \mathbb{R}^{H\times W\times C}$ and $F=\{\textbf{f}_i\}_{i=1}^{L}$ with each $\textbf{f}_i=\Phi(I(i)) \in \mathbb{R}^{H\times W\times C}$, where the $H$ and $W$ represent the spatial resolution of the features and $C$ denotes the channel dimension. For subsequent processing, we flatten $\textbf{q}$ and $F$ to obtain $\textbf{Q}=\mathtt{flatten}(\textbf{q}) \in \mathbb{R}^{HW\times C}$ and $\textbf{V}=\{\textbf{v}_i\}_{i=1}^{L}$ with each $\textbf{v}_i \in \mathbb{R}^{HW\times C}$.

\vspace{0.5em}
\noindent
\textbf{Progressive Knowledge-guided Feature Refinement for EgoVQL.} As mentioned earlier, the core idea of PRVQL is to exploit target knowledge directly from videos and apply it as guidance to enhance query and video features for target localization. For this purpose, PRVQL is implemented as a progressive architecture with multiple stages in a sequence. Each but the last stage involves target knowledge learning and knowledge-guided feature refinement, as in Fig.~\ref{fig:framework}.

More specifically, for the $k^{\text{th}}$ ($1\le k < K$) stage of our PRVQL, let $\mathcal{Q}_{k}$ and $\mathcal{V}_{k}$ denote the query and video features. For the first stage ($k=1$), $\mathcal{Q}_1$ and $\mathcal{V}_1$ are initialized using query and video features extracted from the backbone, and $\mathcal{Q}_{1}=\textbf{Q}$ and $\mathcal{V}_{1}=\textbf{V}$. Otherwise, $\mathcal{Q}_{k}$ and $\mathcal{V}_{k}=\{v_i^k\}_{i=1}^{L}$ are refined features in the last stage $(k-1)$. To mine target-specific knowledge from the video, we perform feature fusion between $\mathcal{Q}_{k}$ and $\mathcal{V}_{k}$, aiming to inject target information into video feature for improving its target awareness. In specific, we leverage cross-attention from~\cite{VaswaniSPUJGKP17} for feature fusion owing to its powerfulness in feature modeling. Mathematically, this process can be expressed as follows,
\begin{equation}\label{eq1}
\setlength{\abovedisplayskip}{5pt} 
\setlength{\belowdisplayskip}{5pt}
\mathcal{X}_k=\{x_i^k | x_i^k = \mathtt{CAB}(v_i^k,\mathcal{Q}_k)\} \;\;\; i=1,2,\cdots,L
\end{equation}
where $\mathcal{X}_k$ is the fused feature in stage $k$, and $v_i^k$ the feature in frame $i$. $\mathtt{CAB}(\mathbf{z},\mathbf{u})$ is the cross-attention (CA) block with $\mathbf{z}$ generating query and $\mathbf{u}$ key/value. Due to space limitation, please see \emph{supplementary material} for detailed architecture. Besides fused feature, we also obtain target-aware spatial attention maps $\mathcal{S}_{k}=\{s_{i}^{k}\}_{i=1}^{L}\in \mathbb{R}^{L\times HW \times HW}$ for $L$ frames in Eq.~(\ref{eq1}), with each $s_{i}^{k}\in \mathbb{R}^{HW \times HW}$ the attention map from the cross-attention operation in $\mathtt{CAB}(v_i^k,\mathcal{Q}_k)$.

To further capture spatio-temporal relations from videos for enhancing features, we apply self-attention~\cite{VaswaniSPUJGKP17} on $\mathcal{X}_k$ by propagating the query information spatially and temporally. Considering that targets in nearby frames are highly correlated, we restrict the attention operation in a temporal window using a masking strategy, similar to~\cite{jiang2024single}. To reduce the computation, we downsample $\mathcal{X}_k$ to decrease the spatial dimension of each frame feature to $h\times w$. Then, we add a position embedding $\mathcal{E}_k^{\text{pos}}$ to the video feature. This process can be written as follows,
\begin{equation}
\setlength{\abovedisplayskip}{6pt} 
\setlength{\belowdisplayskip}{6pt}
    \tilde{\mathcal{X}}_k = \mathtt{Downsample}(\mathcal{X}_k) + \mathcal{E}_k^{\text{pos}}
\end{equation}
where $\mathtt{Downsample}(\cdot)$ represents the downsampling operation implemented with convolution operation. Afterwards, masked self-attention is applied on as $\tilde{\mathcal{X}}$ as follows,
\begin{equation}\label{eq3}
\setlength{\abovedisplayskip}{7pt} 
\setlength{\belowdisplayskip}{7pt}
    \mathcal{H}_k=\mathtt{MaskedSA}(\tilde{\mathcal{X}}_k)
\end{equation}
where $\mathcal{H}_k$ represents enhanced video feature. $\mathtt{MaskedSA}(\mathbf{z})$ denotes the masked self-attention block with $\textbf{z}$ generating query/key/value. In this block, each feature element from frame $i$ only attends to feature elements from frames in the temporal range [($i-u$), ($i+u$)], which can be easily implemented using masking strategy~\cite{VaswaniSPUJGKP17,cheng2022masked}. From Eq.~(\ref{eq3}), besides the $\mathcal{H}_k$, we also gain the temporal-aware spatial attention maps, denoted as $\mathcal{T}_k \in \mathbb{R}^{L\times hw \times Lhw}$, for the target in the video, which will be used for knowledge generation.

\begin{figure}[t]
	\centering
	\includegraphics[width=0.98\linewidth]{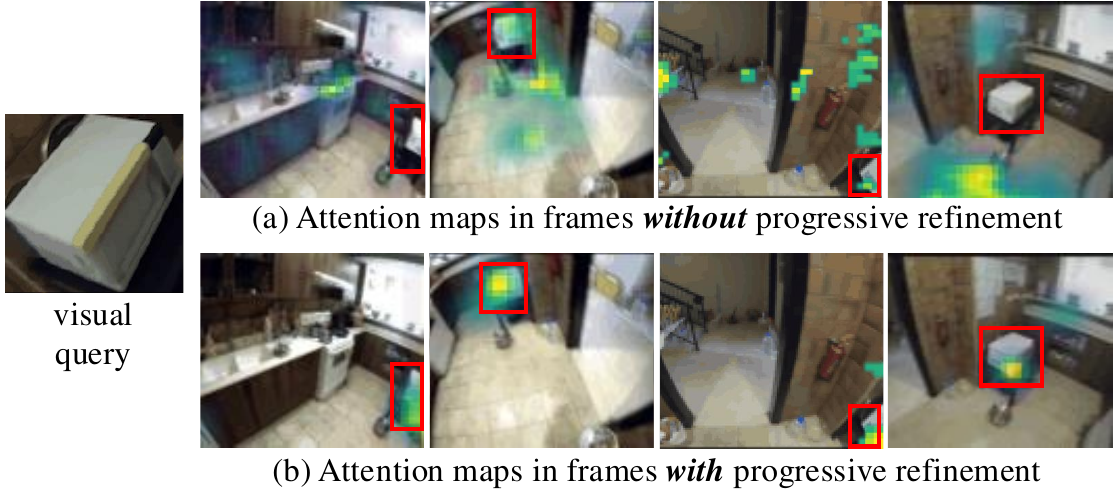}\vspace{-1mm}
	\caption{Comparison of attention maps for video frames from the masked self-attention block \emph{without} (a) and \emph{with} (b) our progressive refinement. As shown, our method can better focus on the target regions, and hence can improve target localization in EgoVQL. The red boxes indicate the foreground object to localize.}
        \vspace{-4mm}
	\label{fig:att}
\end{figure}

With video feature $\mathcal{H}_k$ and attention maps $\mathcal{S}_k$ and $\mathcal{T}_k$, the target knowledge can be extracted with the AKG and SKG modules (as explained later in Sec.~\ref{secakg} and~\ref{skg}), as follows,
\begin{equation}\label{eq_kaks}
\setlength{\abovedisplayskip}{7pt} 
\setlength{\belowdisplayskip}{7pt}
    \mathcal{K}_k^a=\mathtt{AKG}(\mathcal{H}_k, \mathcal{V}_k) \;\;\;\;\;\;\;
    \mathcal{K}_k^s=\mathtt{SKG}(\mathcal{S}_k, \mathcal{T}_k)
\end{equation}
where $\mathcal{K}_k^a$ represents the appearance knowledge and $\mathcal{K}_k^s$ the spatial knowledge. Guided by $\mathcal{K}_k^a$ and $\mathcal{K}_k^s$ in stage $k$, we can refine query and video features using two update modules QFR and VFR (as described later in Sec.~\ref{update}) as follows,
\begin{equation}\label{eq_qfuvfu}
\setlength{\abovedisplayskip}{7pt} 
\setlength{\belowdisplayskip}{7pt}
    \mathcal{Q}_{k+1}=\mathtt{QFR}(\mathcal{K}_k^a, \mathcal{Q}_{k}) \;\;\;\;\;\;\; \mathcal{V}_{k+1}=\mathtt{VFR}(\mathcal{K}_k^s,  \mathcal{V}_1)
\end{equation}
where $\mathcal{Q}_{k+1}$ and $\mathcal{V}_{k+1}$ are refined features guided by target knowledge, which are fed to the next stage $(k+1)$ to generate more accurate knowledge for further feature refinement. Fig.~\ref{fig:att} compares the attention maps from the masked self-attention with and without using our approach. We can see that, our method with refined features guided by knowledge can better focus on the target in the video and thus improves target localization, showing its efficacy.

For the final $K^{\text{th}}$ stage in PRVQL, since no knowledge is extracted, the AKG and SKG modules are removed. Given the visual query and video features $\mathcal{Q}_{K}$ and $\mathcal{V}_{K}$ from the $(K-1)^{\text{th}}$ stage, we can then obtain the final enhanced video feature $\mathcal{H}_{K}$ through Eqs.~(\ref{eq1})-(\ref{eq3}) in the $K^{\text{th}}$ stage. With $\mathcal{H}_{K}$, we use the prediction heads as in ~\cite{jiang2024single} for target localization via regression and classification. For details of the adopted prediction heads, please kindly refer to~\cite{jiang2024single}.

\subsection{Appearance Knowledge Generation (AKG)}\label{secakg}

In order to extract discriminative visual information of target directly from the video, we introduce a simple yet highly effective module, named \emph{appearance knowledge generation} (AKG), for appearance knowledge learning. Specifically, it first estimates the potential target regions from the video using target-aware video features. Then, based on confidence scores of these regions, we select the top few ones to extract target features from the video as the appearance knowledge. 

Specifically, given the target-aware video feature $\mathcal{H}_k$, we first reshape it to the 2D feature map, and then increase its spatial resolution back to $H\times W$ as follows,
\begin{equation}
\setlength{\abovedisplayskip}{6pt} 
\setlength{\belowdisplayskip}{6pt}
    \tilde{\mathcal{H}}_k=\mathtt{Upsample}(\mathtt{Reshape}(\mathcal{H}_k))
\end{equation}
where $\mathtt{Upsample}(\cdot)$ denotes the upsampling operation. After this, we apply $\tilde{\mathcal{H}}_k$ to produce temporal confidence scores and spatial box regions for target in each frame. More concretely, we first split $\tilde{\mathcal{H}}_k$ along the channel dimension into two equal halves $\tilde{\mathcal{H}}_k^{t}$ and $\tilde{\mathcal{H}}_k^{s}$ via $\tilde{\mathcal{H}}_k^{t}, \tilde{\mathcal{H}}_k^{s}=\mathtt{Split}(\tilde{\mathcal{H}}_k)$. Inspired by~\cite{jiang2024single}, we perform classification and regression to predict temporal confidence scores and spatial boxes using multi-scale anchors~\cite{RenHGS15}. Specifically, two Conv blocks are applied on $\tilde{\mathcal{H}}_k^{t}(i)$ and $\tilde{\mathcal{H}}_k^{s}(i)$ for prediction as follows,
\begin{equation}\label{akg}
\setlength{\abovedisplayskip}{7pt}
\setlength{\belowdisplayskip}{7pt}
    \tilde{\mathcal{C}}_k=\mathtt{ConvBlock}(\tilde{\mathcal{H}}_k^{t}) \;\;\;\;\; \Delta\tilde{\mathcal{B}}_k=\mathtt{ConvBlock}(\tilde{\mathcal{H}}_k^{s})
\end{equation}
where $\tilde{\mathcal{C}}_k \in \mathbb{R}^{L \times H \times W \times m}$ denotes the temporal confidence scores for target in $L$ frames with $m$ the number of anchors at each position. $\Delta\tilde{\mathcal{B}}_k \in \mathbb{R}^{L \times H \times W \times 4m}$ is the offsets to the anchor boxes $\bar{\mathcal{B}}$, and target boxes $\tilde{\mathcal{B}}_k=\Delta\tilde{\mathcal{B}}_k + \bar{\mathcal{B}}$. With $\tilde{\mathcal{C}}_k$, the confidence score in each frame is determined by the highest classification score of anchors, and the target region is the box corresponding to the box with the highest classification score. This way, we can obtain the confidence scores $\bar{\mathcal{C}}_k$ and target regions $\bar{\mathcal{B}}_k$ in each frames as follows, 
\begin{equation}\label{pred}
\setlength{\abovedisplayskip}{7pt}
\setlength{\belowdisplayskip}{7pt}
\begin{split}
    &\bar{\mathcal{C}}_k = \{c_{k}^{i} | c_{k}^{i}, d_k^i = \mathtt{Max}(\tilde{\mathcal{C}}_k(i))\} 
    \\
    &\bar{\mathcal{B}}_k = \{b_{k}^{i} | b_{k}^{i} = \mathtt{Index}(\tilde{\mathcal{B}}_k(i), d_k^i)\} 
\end{split}
\end{equation}
where $i\in[1,L]$ is the frame index. $c_{k}^{i}$ is the highest value selected from the classification scores $\tilde{\mathcal{C}}_k(i)\in\mathbb{R}^{H\times W\times m}$ of anchors in frame $i$, and $d_k^i$ is its index. $b_{k}^{i}$ is the target box corresponding to $c_{k}^{i}$ in frame, and extracted from $\tilde{\mathcal{B}}_k(i)\in\mathbb{R}^{H\times W\times 4m}$. $\mathtt{Max}(\cdot)$ is to select the maximum and its index, and $\mathtt{Index}(\cdot)$ to extract the box from $\tilde{\mathcal{B}}_k(i)$ given its index. 

\begin{figure}[t]
	\centering
	\includegraphics[width=\linewidth]{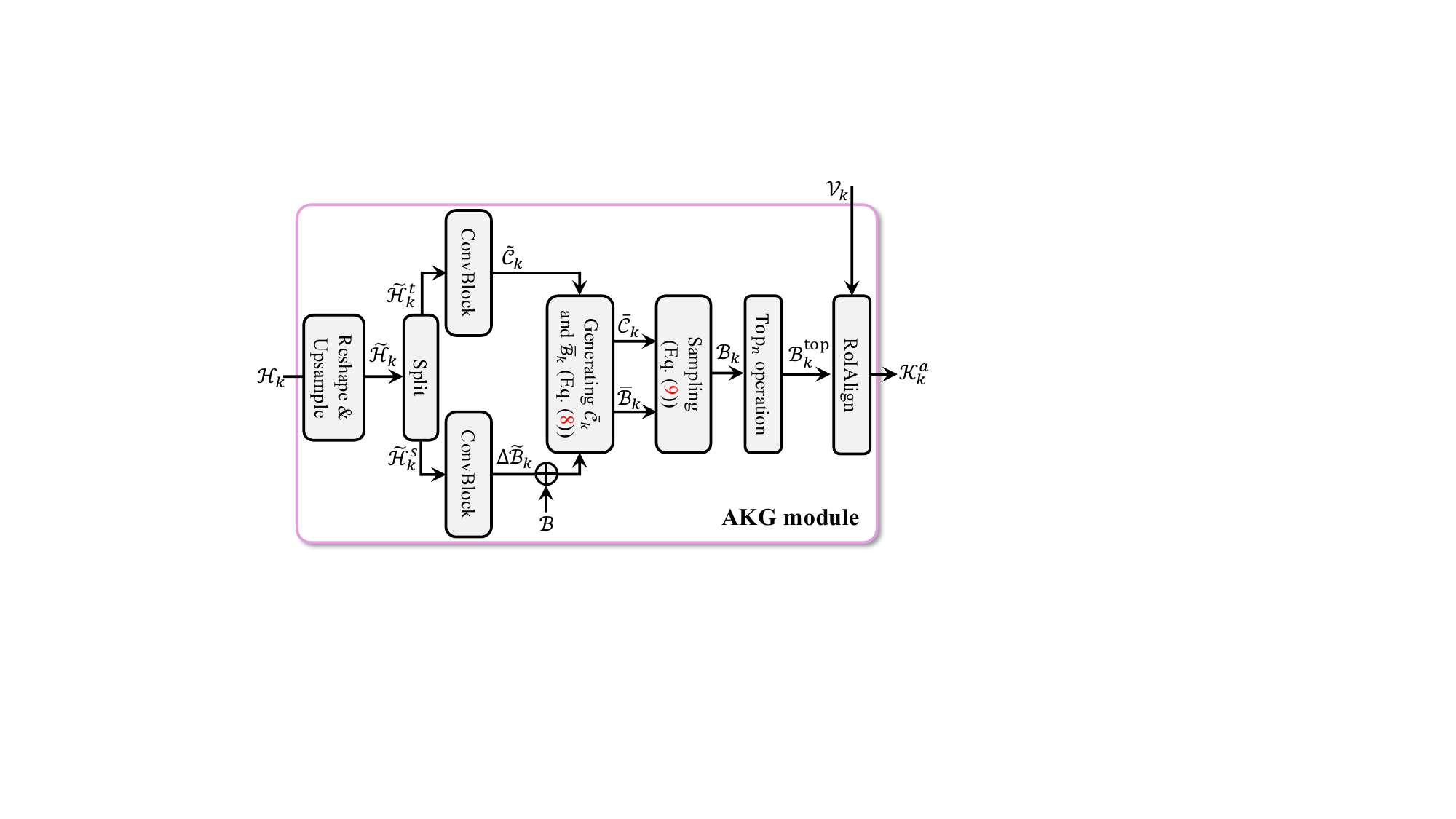}\vspace{-1mm}
	\caption{Illustration of appearance knowledge generation (AKG).}
        \vspace{-3mm}
	\label{fig:akgfig}
\end{figure}

With $\bar{\mathcal{C}}_k\in\mathbb{R}^{L\times 1}$ and $\bar{\mathcal{B}}_k\in\mathbb{R}^{L\times 4}$, we first sample target regions with high confidence scores as follows,
\begin{equation}\label{threshold}
\setlength{\abovedisplayskip}{7pt}
\setlength{\belowdisplayskip}{7pt}
\mathcal{B}_k=\mathtt{Sample}(\bar{\mathcal{B}}_k(i), \bar{\mathcal{C}}_k(i), \tau) =\{\bar{\mathcal{B}}_k(i) | \bar{\mathcal{C}}_k(i) > \tau\}
\end{equation}
Then, we extract $n$ regions from $\mathcal{B}_k$ with the top confidence scores via $\mathcal{B}_k^{\text{top}}=\mathtt{Top}_{n}(\mathcal{B}_k)$. If the number of regions in $\mathcal{B}_k$ is less then $n$, we keep all regions. After this, RoIAlign~\cite{he2017mask} is used to extract the appearance knowledge from $\mathcal{V}_{k}$ via
\begin{equation}\label{roi}
\setlength{\abovedisplayskip}{7pt}
\setlength{\belowdisplayskip}{7pt}
\mathcal{K}_{k}^{a}=\mathtt{RoIAlign}(\mathcal{V}_{k}, \mathcal{B}_k^{\text{top}})
\end{equation}
where $\mathcal{K}_{k}^{a}$ represents the appearance knowledge from AKG in the $k^{\text{th}}$ stage. Please notice that, in Eq.~(\ref{roi}), we only perform RoIAlign in frames corresponding to $\mathcal{B}_k^{\text{top}}$. Since $\mathcal{K}_{k}^{a}$ is generated from the video itself, when using it as guidance to refine the query feature, we can reduce the discrepancy between the query and the foreground target. By deploying AKG in each but the last stage, $\mathcal{K}_{k}^{a}$ could be gradually improved with better refined query feature in each stage. Fig.~\ref{fig:akgfig} illustrates AKG for appearance knowledge generation.

\subsection{Spatial Knowledge Generation (SKG)}\label{skg}

In addition to appearance knowledge, we explore target spatial knowledge from the video for improving video features. Specifically, inspired by the \emph{observation} that intermediate attention maps from previous attention operations reflect the spatial cues of target in each frame to some extent, similar to the concept of ``\emph{saliency}'' but for the target, we propose the \emph{spatial knowledge generation} (SKG) module, which works to leverage readily available attention maps as guidance for enhancing target while suppressing background in the video features, enabling more focus on the target in PRVQL.

\begin{figure}[t]
	\centering
	\includegraphics[width=0.9\linewidth]{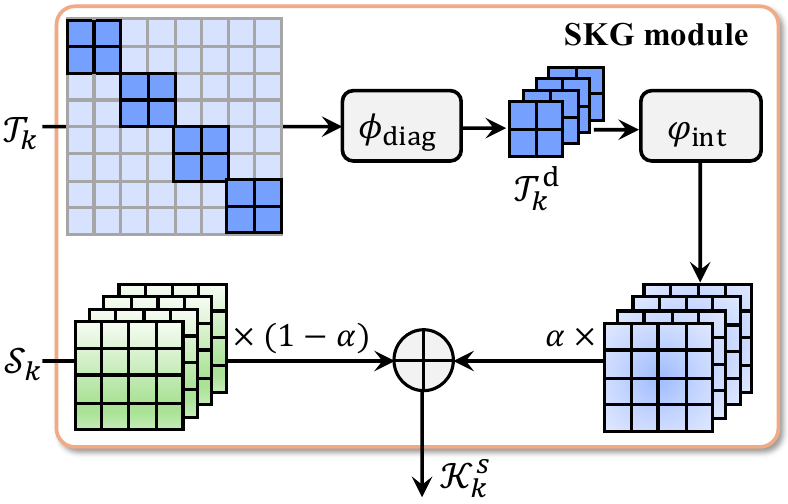}\vspace{-1mm}
	\caption{Illustration of spatial knowledge generation (SKG).}
        \vspace{-3mm}
	\label{fig:skgfig}
\end{figure}

Concretely in our SKG, we exploit the target-aware spatial attention maps $\mathcal{S}_k$ from cross-attention block in Eq.~(\ref{eq1}) and temporal-aware spatial attention maps $\mathcal{T}_k$ from masked self-attention block in Eq.~(\ref{eq3}) for spatial knowledge learning. Specifically, given $\mathcal{S}_k$ and $\mathcal{T}_k$, we first extract the inter-frame spatial attention maps $\mathcal{T}_k^{\text{d}}$ by extracting diagonal elements from $\mathcal{T}_k$ as follows,
\begin{equation}\label{inter-frame}
\setlength{\abovedisplayskip}{7pt}
\setlength{\belowdisplayskip}{7pt}
\mathcal{T}_k^{\text{d}}=\phi_{\text{diag}}(\mathcal{T}_k)=\{t_{i}^k\}_{i=1}^{L}
\end{equation}
where $\phi_{\text{diag}}$ denotes the operation to extract diagonal elects, and $t_{i}^k\in \mathbb{R}^{hw\times hw}$ represents the attention maps for frame $i$. To match the spatial dimension of $\mathcal{T}_k^{\text{d}}$ and $\mathcal{S}_k$, we first perform bilinear interpolation on $\mathcal{T}_k^{\text{d}}$ to increase its spatial resolution to $HW \times HW$, and then combines these two attention maps to obtain spatial knowledge. Mathematically, this process can be expressed as follows,
\begin{equation}\label{inter-frame2}
\setlength{\abovedisplayskip}{7pt}
\setlength{\belowdisplayskip}{7pt}
\mathcal{K}_k^{s}=\alpha \cdot \varphi_{\text{int}}(\mathcal{T}_k^{\text{d}}) + (1-\alpha) \cdot \mathcal{S}_k
\end{equation}
where $\varphi_{\text{int}}$ denotes the bilinear interpolation operation, $\mathcal{K}_k^{s}$ is the target spatial knowledge, and $\alpha$ is a parameter to balance different attention maps. Since $\mathcal{K}_k^{s}$ indicates the target position cues in each frame in some degree, we can use it to highlight target while restraining background in videos for improving localization. Similar to AKG, SKG is deployed in each but the last stage of PRVQL. Fig.~\ref{fig:skgfig} illustrates SKG. 

\subsection{Feature Refinement with Knowledge}\label{update}

With target appearance knowledge $\mathcal{K}_{k}^{a}$ and spatial knowledge $\mathcal{K}_{k}^{s}$ obtained from AKG and SKG in stage $k$ ($1<k\le K$), we then apply them as guidance to refine the query and video features through \emph{query feature refinement} (QFR) and \emph{video feature refinement} (VFR) modules.

\vspace{0.3em}
\noindent
\textbf{Query Feature Refinement (QFR).} QFR aims to refine the query feature with guidance from learned target appearance knowledge. Specifically, it adopts a cross-attention block to fuse appearance knowledge $\mathcal{K}_{k}^{a}$ into the query. Given the query feature $\mathcal{Q}_k$ and appearance knowledge $\mathcal{K}_{k}^{a}$ in stage $k$, we first apply a Conv block on $\mathcal{K}_{k}^{a}$ and then perform refinement via QFR as follows,
\begin{equation}\label{eq:q_update}
\setlength{\abovedisplayskip}{7pt} 
\setlength{\belowdisplayskip}{7pt}
\mathcal{Q}_{k+1} = \mathtt{QFR}(\mathcal{Q}_{k}, \mathcal{K}_{k}^{a}) = \mathtt{CAB}(\mathcal{Q}_{k}, \mathtt{CNB}(\mathcal{K}_k^a))
\end{equation}
where $\mathcal{Q}_{k+1}$ is refined query feature and fed to next stage for learning more accurate knowledge, which in turn leads to better query feature for localization in the final stage. It is worth noting that, besides cross-attention, we explore different strategies to combine $\mathcal{Q}_k$ and $\mathcal{K}_{k}^{a}$, including addition and concatenation operations. We observe that using cross-attention achieves the best performance, as exhibited in our experiments provided in the \emph{supplementary material}.

\vspace{0.3em}
\noindent
\textbf{Video Feature Refinement (VFR).} VFR focuses on adopting target spatial knowledge to refine initial video feature by enhancing target while suppressing the background regions. Concretely, given the initial video feature $\mathcal{V}_1$ and learn spatial knowledge $\mathcal{K}_{k}^{s}$ in stage $k$, we use residual connection to refine $\mathcal{V}_1$ as follows,
\begin{equation}\label{eq:v_update}
\setlength{\abovedisplayskip}{7pt} 
\setlength{\belowdisplayskip}{7pt}
\mathcal{V}_{k+1} = \beta \cdot (\mathcal{K}_{k}^{s} \odot \mathcal{V}_1) + (1-\beta) \cdot \mathcal{V}_1
\end{equation}
where $\mathcal{V}_{k+1}$ denotes the refined video feature that is used for the next stage, $\beta$ is a balancing parameter, and $\odot$ represents the pixel-wise multiplication.

\subsection{Optimization and Inference}\label{opt}

\textbf{Optimization}. Given a video and a visual query, we predict confidence scores $\tilde{\mathcal{C}}_k$ and target boxes $\tilde{\mathcal{B}}_k$ ($\tilde{\mathcal{B}}_k=\Delta\tilde{\mathcal{B}}_k+\tilde{\mathcal{B}}$) in each stage $k$ ($1\le k\le K$). During training, given the groundtruth boxes $\mathcal{B}^*$ and temporal occurrence scores $\mathcal{S}^*$, we design the following loss function $\mathcal{L}_k$ for stage $k$,
\begin{equation}\label{eq:loss}
\setlength{\abovedisplayskip}{7pt} 
\setlength{\belowdisplayskip}{7pt}
\mathcal{L}_k = \mathcal{L}_{\text{L}_1}(\tilde{\mathcal{B}}_k, \mathcal{B}^*) + \lambda_1\mathcal{L}_{\text{GIoU}}(\tilde{\mathcal{B}}_k, \mathcal{B}^*) + \lambda_2\mathcal{L}_{\text{BCE}}(\tilde{\mathcal{S}}_k, \mathcal{S}^*)
\end{equation}
where $\mathcal{L}_{\text{L}_1}$, $\mathcal{L}_{\text{GIoU}}$, and $\mathcal{L}_{\text{BCE}}$ represent the $L_1$ loss, generalized IoU (GIoU)~\cite{rezatofighi2019generalized} loss, and binary cross-entropy (BCE) loss, respectively. $\lambda_1$ and $\lambda_2$ are two balancing parameters. With Eq.~(\ref{eq:loss}), the total training loss $\mathcal{L}_{\text{total}}$ can be obtained via $\mathcal{L}_{\text{total}}=\sum_{k=1}^{K} \mathcal{L}_k$. Following~\cite{jiang2024single,xu2023my,grauman2022ego4d}, we perform hard negative mining during training to decrease false positive prediction. For details, please refer to~\cite{jiang2024single,xu2023my,grauman2022ego4d}.

\vspace{0.3em}
\noindent
\textbf{Inference.} We employ the same strategy as in~\cite{jiang2024single} to obtain the prediction result. Specifically, during inference, we first obtain the target region in each frame by selecting target box with the highest confidence score. Please note that, the target regions with confidences scores smaller than a threshold, set to 0.79, will be discarded. After this, we select the most recent peak and generate a response track via bidirectional search from the peak. Details can be seen in~\cite{jiang2024single}.

\section{Experiments}

\textbf{Implementation.} Our PRVQL is implemented using PyTorch~\cite{paszke2019pytorch} with Nvidia RTX A6000 GPUs. Similar to~\cite{jiang2024single}, we use the popular ViT~\cite{DosovitskiyB0WZ21} pretrained with DINOv2~\cite{OquabDMVSKFHMEA24} as the backbone. Our PRVQL is end-to-end trained for 50 epoches (a total of 60K iterations) with a batch size of 12, utilizing the AdamW optimizer~\cite{LoshchilovH19} with a peak learning rate of $10^{-4}$ and a weight decay of $5\times10^{-2}$. The query image and video frames are resized to $480\times480$. The number of stages $K$ in PRVQL is empirically set to 3, and the pooling size for RoIAlign is 5. The number of selected boxes $n$ for appearance knowledge is 3, and the threshold $\tau$ is set to $0.7$. The parameter $\alpha$ for computing spatial knowledge is empirically set to 0.5. The balancing parameter $\beta$ is 0.1. $\lambda_1$ and $\lambda_2$ are empirically set to 0.3 and 100. The video frame length $L$, similar to~\cite{jiang2024single}, is set to $30$ with frames randomly selected to ensure coverage of at least a portion of the response track. For the anchor boxes in localization, we employ four scales ($16^{2}$, $32^{2}$, $64^{2}$, $48^{2}$) with three aspect ratios (0.5, 1, 2) for each anchor box, similar to~\cite{jiang2024single}.

\begin{table}[!t]
	\centering
        \caption{Comparison on the Ego4D validation set.}\vspace{-2mm}
	\renewcommand{\arraystretch}{1.05}
	\scalebox{0.92}{
	\begin{tabular}{rcccc}
	\rowcolor{mygray}
	\specialrule{1.5pt}{0pt}{0pt}
	Methods & tAP$_{25}$ & stAP$_{25}$ & rec\% &  Succ    \\
	\hline
	\hline
        STARK \textcolor{lightblue}{\scriptsize{[ICCV'21]}}  & 0.10   & 0.04   & 12.41  & 18.70    \\
	SiamRCNN \textcolor{lightblue}{\scriptsize{[CVPR'22]}}  & 
        0.22   & 0.15   & 32.92  & 43.24    \\
	NFM \textcolor{lightblue}{\scriptsize{[VQ2D Challenge'22]}}     & 0.26  & 0.19  & 37.88  & 47.90    \\
	CocoFormer \textcolor{lightblue}{\scriptsize{[CVPR'23]}} 
        & 0.26  & 0.19  & 37.67  & 47.68    \\
	VQLoC \textcolor{lightblue}{\scriptsize{[NeurIPS'23]}}  
        & 0.31  & 0.22  & 47.05  & 55.89    \\
        \hline
	\rowcolor{highlight} PRVQL (ours) & \textbf{0.35} & \textbf{0.27} & 
        \textbf{47.87} & \textbf{57.93}   \\
        \specialrule{1.5pt}{0pt}{0pt}
	\end{tabular}}
        \label{tab:sota_val}
\end{table}

\begin{table}[!t]
	\centering
        \caption{Comparison on the Ego4D test set.}\vspace{-2mm}
	\renewcommand{\arraystretch}{1.05}
	\scalebox{0.92}{
	\begin{tabular}{rcccc}
	\rowcolor{mygray}
	\specialrule{1.5pt}{0pt}{0pt}
	Methods & tAP$_{25}$ & stAP$_{25}$ & rec\% &  Succ    \\
	\hline
	\hline
        STARK \textcolor{lightblue}{\scriptsize{[ICCV'21]}}  & -   & -   & -  & -    \\
	SiamRCNN \textcolor{lightblue}{\scriptsize{[CVPR'22]}}  & 
        0.20   & 0.13   & -  & -    \\
	NFM \textcolor{lightblue}{\scriptsize{[VQ2D Challenge'22]}}     & 0.24  & 0.17  & -  & -    \\
	CocoFormer \textcolor{lightblue}{\scriptsize{[CVPR'23]}} 
        & 0.25  & 0.18  & -  & -    \\
	VQLoC \textcolor{lightblue}{\scriptsize{[NeurIPS'23]}}  
        & 0.32  & 0.24  & 45.11  & 55.88    \\
        \hline
	\rowcolor{highlight} PRVQL (ours) & \textbf{0.37} & \textbf{0.28} & 
        \textbf{45.70} & \textbf{59.43}   \\
        \specialrule{1.5pt}{0pt}{0pt}
	\end{tabular}}
        \label{tab:sota_tst}
\end{table}

\begin{table}[!t]\small
\setlength{\tabcolsep}{2.2pt}
	\centering
        \caption{Comparison of speed on Ego4D.}\vspace{-2mm}
	\renewcommand{\arraystretch}{1.05}
	\scalebox{0.95}{
	\begin{tabular}{rcccccc}
	\rowcolor{mygray}
	\specialrule{1.5pt}{0pt}{0pt}
	 & STARK & SiamRCNN & NFM &  CocoFormer & VQLoC & PRVQL    \\
	\hline
	\hline
        FPS  & 33   & 3   & 3  & 3 & 36 &  30  \\
        \specialrule{1.5pt}{0pt}{0pt}
	\end{tabular}}
        \label{tab:sota_fps}\vspace{-3mm}
\end{table}

\subsection{Dataset and Evaluation Metrics}

\textbf{Dataset.} Following previous methods~\cite{xu2023my,jiang2024single}, we conduct experiments on the challenging Ego4D benchmark~\cite{grauman2022ego4d}. Ego4D is a recently proposed large-scale dataset dedicated to first-person video understanding. Similar to~\cite{jiang2024single}, we use videos from the VQ2D task. There are 13.6K, 4.5K, 4.4K pairs of queries and videos for training, validation, and testing, lasting 262, 87, and 84 hours, respectively.

\vspace{0.5em}
\noindent
\textbf{Evaluation Metrics.} Following~\cite{xu2023my,jiang2024single}, we adopt the metrics provided by Ego4D~\cite{grauman2022ego4d} for evaluation, including temporal average precision (tAP$_{25}$), spatio-temporal average precision (stAP$_{25}$), recovery (rec\%), and success (Succ). tAP$_{25}$ and stAP$_{25}$ are used to measure the accuracy of the predicted temporal and spatio-temporal extends of the of response tracks in comparison to groundtruth using the Intersection over Union (IoU) with threshold 0.25. The recovery metric assess the percentage of predicted frames in which the IoU between predicted bounding box and ground-truth is great than or equal to 0.5, and success metric measures weather the IoU between prediction and groundtruth exceeds 0.05. For more details of metrics, please refer to~\cite{grauman2022ego4d}.

\subsection{State-of-the-art Comparison}

In order to verify the effectiveness of our PRVQL, we compare it with other state-of-the-art methods on Ego4D, including STARK~\cite{yan2021learning}, SiamRCNN~\cite{voigtlaender2020siam}, NFM~\cite{xu2022negative}, CocoFormer~\cite{xu2023my}, and VQLoC~\cite{jiang2024single}. Tab.~\ref{tab:sota_val} displays the results and comparison on the Ego4D validate test. As in Tab.~\ref{tab:sota_val}, we can clearly see that the proposed PRVQL achieves the best performance on all four metrics. In particular, it achieves the 0.35 tAP$_{25}$ and 0.27 stAP$_{25}$ scores, which outperforms the second best method VQLoC with 0.31 tAP$_{25}$ and 0.22 stAP$_{25}$ scores by 4\% and 5\%. Besides, the rec and Succ scores of PRVQL are 47.87\% and 57.93 respectively, which surpasses the 47.05\% rec\ and 55.89 Succ scores of VQLoC, evidencing the effectiveness of our approach. In addition, in Tab.~\ref{tab:sota_tst} we further report the experimental results and comparison on Ego4D test set. As in Tab.~\ref{tab:sota_tst}, our PRVQL again achieves the best performance on all four metrics. Specifically, PRVQL obtains the 0.37 tAP$_{25}$ and 0.28 stAP$_{25}$ scores. Compared to the second best method VQLoC, our approach outperforms it by 5\% and and 4\%, respectively, on tAP$_{25}$ and stAP$_{25}$. In addition, the rec and Succ scores of PRVQL are 45.70\% and 59.43, which are better than those of VQLoC with 45.11\% and 55.88. All these show the efficacy of target knowledge in improving EgoVQL.

In addition, we show the comparison of speed, measured by frames per second (\emph{FPS}), for different methods in Tab.~\ref{tab:sota_fps}. From Tab.~\ref{tab:sota_fps}, we can see our method runs fast at a speed of 30 FPS. Despite being slightly slower than VQLoC running at a speed of 36 FPS, our PRVQL is more robust in localization, showing a better balance between accuracy and speed.

\begin{table}[!t]
	\centering
        \caption{Ablation studies of AKG and SKG.}\vspace{-2mm}
	\renewcommand{\arraystretch}{1.1}
	\scalebox{0.92}{
		\begin{tabular}{cccccccc}
			\specialrule{1.5pt}{0pt}{0pt}
			\rowcolor{mygray} 
			& AKG & SKG & tAP$_{25}$ & stAP$_{25}$ & rec\% &  Succ \\ \hline\hline
			\ding{182} & - & - & 0.32 & 0.23 & 45.24 & 55.37 \\
			\ding{183} & \checkmark & - & 0.34 & 0.26 & 47.34 & 57.27 \\
            \ding{184} & - & \checkmark & 0.33 & 0.24 & 46.33 & 56.46 \\
			\ding{185} & \checkmark & \checkmark & \textbf{0.35} & \textbf{0.27} & \textbf{47.87} & \textbf{57.93} \\ \specialrule{1.5pt}{0pt}{0pt}
	\end{tabular}}
	\label{KG_modules}
\end{table}

\begin{table}[!t]
        \setlength{\tabcolsep}{7.7pt}
	\centering
        \caption{Ablation studies on the number of stages.}\vspace{-2mm}
	\renewcommand{\arraystretch}{1}
	\scalebox{0.92}{
		\begin{tabular}{cccccc}
		\specialrule{1.5pt}{0pt}{0pt}
		\rowcolor{mygray} 
		& \# Stages & tAP$_{25}$ & stAP$_{25}$ & rec\% &  Succ \\ \hline\hline
           \ding{182} & $K=1$ & 0.32 & 0.23 & 45.24 & 55.37 \\
           \ding{183} & $K=2$ & 0.34 & \textbf{0.27} & 47.25 & 56.43 \\
           \ding{184} & $K=3$ & \textbf{0.35} & \textbf{0.27} & \textbf{47.87} & \textbf{57.93} \\
           \ding{185} & $K=4$ & 0.33 & 0.26 & 45.91 & 55.29 \\
	\specialrule{1.5pt}{0pt}{0pt}
    \end{tabular}}
	\label{tab:stage}\vspace{-3mm}
\end{table}

\subsection{Ablation Study}

For better understanding of PRVQL, we conduct extensive ablation studies on Ego4D validation set as follows. 

\vspace{0.3em}
\noindent
\textbf{Impact of AKG and SKG.} AKG and SKG are two important modules in PRVQL for target appearance and spatial knowledge generation. In order to analyze these two modules, we conduct thorough ablation studies in Tab.~\ref{KG_modules}. From Tab.~\ref{KG_modules}, we can see that, without AKG and SKG, the tAP$_{25}$ and stAP$_{25}$ scores are 0.32 and 0.23, respectively (\ding{182}). By applying AKG alone for refinement with appearance knowledge, they can be significantly improved to 0.34 and 0.26 with performance gains of 0.02 and 0.03 (\ding{183} \emph{v.s.} \ding{182}). When using only SKG for refinement with spatial knowledge, tAP$_{25}$ and stAP$_{25}$ are improved to 0.33 and 0.24 (\ding{184} \emph{v.s.} \ding{182}). From this table, we can also observe that, using appearance knowledge for refinement in PRVQL brings more gains than the spatial knowledge (\ding{183} \emph{v.s.} \ding{184}). When using both AKG and SKG in our PRVQL, we achieve the best performance with 0.35 tAP$_{25}$ and 0.27 stAP$_{25}$ scores (\ding{185} \emph{v.s.} \ding{182}), which clearly evidences the efficacy of target knowledge for improving the robustness of EgoVQL.

\begin{table}[!t]
        \setlength{\tabcolsep}{7.5pt}
	\centering
    \caption{Ablation studies on the threshold $\tau$.}\vspace{-2mm}
			\renewcommand{\arraystretch}{1}
			\scalebox{0.92}{
				\begin{tabular}{ccccccc}
					\specialrule{1.5pt}{0pt}{0pt}
					\rowcolor{mygray} 
					& Threshold & tAP$_{25}$ & stAP$_{25}$ & rec\% &  Succ \\ \hline\hline
					\ding{182}& $\tau=0.6$ & 0.32 & 0.24 & 44.82 & 55.97 \\
                      \ding{183} &  $\tau=0.7$ & \textbf{0.35} & \textbf{0.27} & \textbf{47.87} & \textbf{57.93} \\
                       \ding{184} & $\tau=0.8$ & 0.34 & 0.26 & 46.53 & 57.33 \\
					\specialrule{1.5pt}{0pt}{0pt}
			\end{tabular}}
			\label{top_threshold}
\end{table}

\begin{table}[!t]
        \setlength{\tabcolsep}{8.8pt}
	\centering
    \caption{Ablation studies on the number of target boxes in AKG.}\vspace{-2mm}   
	\renewcommand{\arraystretch}{1}
			\scalebox{0.92}{
				\begin{tabular}{cccccc}
					\specialrule{1.5pt}{0pt}{0pt}
					\rowcolor{mygray} 
					 & & tAP$_{25}$ & stAP$_{25}$ & rec\% &  Succ \\ \hline\hline
                      \ding{182} &  $n=2$ & 0.34 & 0.26 & 47.27 & 56.03 \\
                       \ding{183} & $n=3$  & 0.35 & \textbf{0.27} & \textbf{47.87} & \textbf{57.93} \\
                       \ding{184} & $n=4$  & \textbf{0.36} & 0.26 & 46.59 & 57.62 \\
                       \ding{185} & $n=5$  & 0.35 & 0.24 & 46.84 & 56.95 \\
					\specialrule{1.5pt}{0pt}{0pt}
			\end{tabular}}
			\label{tab:top_k}
\end{table}

\begin{table}[!t]
    \setlength{\tabcolsep}{8.6pt}
    \centering
    \caption{Ablation studies on RoIAlign feature size.}\vspace{-2mm}
    \renewcommand{\arraystretch}{1}
    \scalebox{0.92}{
	\begin{tabular}{ccccccc}
	\specialrule{1.5pt}{0pt}{0pt}
	\rowcolor{mygray} 
					& Size & tAP$_{25}$ & stAP$_{25}$ & rec\% &  Succ \\ \hline\hline
				  \ding{182} & 3 & 0.33 & 0.26 & 46.94 & 56.06 \\
                    \ding{183} & 5 & \textbf{0.35} & \textbf{0.27} & \textbf{47.87} & \textbf{57.93} \\
                    \ding{184} & 7 & 0.34 & \textbf{0.27} & 47.58 & 57.38 \\
                    \ding{185} & 9 & 0.32 & 0.25 & 46.37 & 55.27 \\
					\specialrule{1.5pt}{0pt}{0pt}
			\end{tabular}}
			\label{tab:roisize}\vspace{-3mm}
\end{table}

\begin{figure*}[!t]
    \centering
    \includegraphics[width=0.9\linewidth]{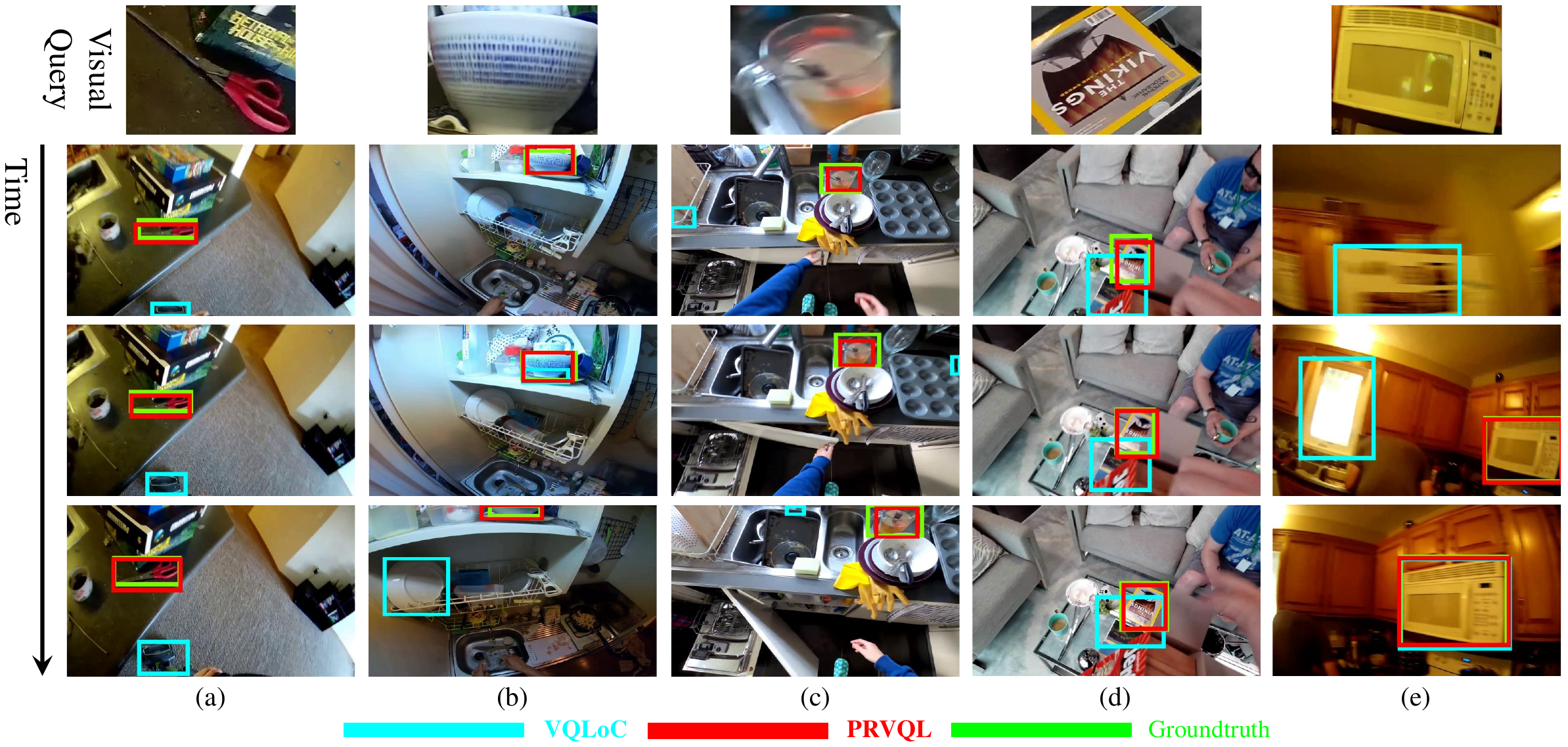}\vspace{-2mm}
    \caption{Qualitative analysis and comparison between our PRVQL and state-of-the-art VQLoC in representative videos with different challenges. We observe that, owing to our target knowledge from videos, PRVQL can more robustly localize the target of interest.}
    \label{fig:visual_comparison}\vspace{-3mm}
\end{figure*}

\vspace{0.3em}
\noindent
\textbf{Impact of the number of stages.} Our PRVQL is designed as a progressive architecture with $K$ stages to explore the target knowledge for refinement. In this work, we conduct an ablation study on the number of stages $K$ in PRVQL as shown in Tab.~\ref{tab:stage}. From Tab.~\ref{tab:stage}, we observe that, when setting $K=1$, which means only one stage is used and the target knowledge is not used due to one-stage design, the tAP$_{25}$ and stAP$_{25}$ scores are 0.32 and 0.23 (\ding{182}). When adding the second stage, tAP$_{25}$ and stAP$_{25}$ can be largely improved by 2\% and 4\% to 0.34 and 0.27, respectively (\ding{183}). With three stages, the tAP$_{25}$ and stAP$_{25}$ scores can be further boosted to 0.35 and 0.27 (\ding{184}). When setting $K=4$ with 4 stages, the performance is decreased with 0.33 tAP$_{25}$ and 0.26 stAP$_{25}$ scores (\ding{185}). Therefore, we set $K$ to 3 in this work.

\vspace{0.3em}
\noindent
\textbf{Impact of threshold $\tau$ in AKG.} The threshold $\tau$ is used to filter out less confident target regions in AKG, aiming to avoid noisy features in appearance knowledge generation. In this work, we conduct an ablation to study the impact of $\tau$ on the final performance in Tab.~\ref{top_threshold}. As shown in Tab.~\ref{top_threshold}, we can see that, when setting $\tau$ to 0.7, PRVQL achieves the best performance on all four metrics (\ding{183}). 

\vspace{0.3em}
\noindent
\textbf{Impact of number of target boxes in AKG.} In AKG, we extract visual features from the top $n$ highly confident target regions for appearance knowledge generation. We conduct an ablation on $n$ in Tab.~\ref{tab:top_k}. From Tab.~\ref{tab:top_k}, we can observe that, when using the top 3 target regions for knowledge learning in AKG, we achieve the best overall performance (\ding{183}).

\vspace{0.3em}
\noindent
\textbf{Impact of RoIAlign Feature Size.} With the top $n$ selected target regions, we perform the RoIAlign operation~\cite{he2017mask} to obtain target appearance knowledge. The  RoIAlign feature size may have an impact on the target appearance knowledge. A too small size may result in the coarse spatial information of the target, while a too large size may lead to losing discriminative local features for the target, both degrading performance. In this work, we study different RoIAlign feature sizes in Tab.~\ref{tab:roisize}. As shown, when setting the size to 5 in RoIAlign, PRVQL shows the best overall performance.

\subsection{Qualitative Analysis}

In order to provide further analysis of our PRVQL, we show the visualization results of its localization and compare it with the state-of-the-art VQLoC in Fig.~\ref{fig:visual_comparison}. Specifically, we show the results and comparison on several representative videos, including video in (a) with \emph{pose variation}, video in (b) with \emph{cluttering background} and \emph{out-of-view}, video in (c) with \emph{occlusion} and \emph{low resolution}, video in (d) with \emph{pose variation} and \emph{cluttering background}, and video in (d) with \emph{motion blur} and \emph{distractor}. From Fig.~\ref{fig:visual_comparison}, we can observe that, our method can robustly and accurately localize the target of interest in all these challenges, owing to the help of target knowledge from the videos, while VQLoC is prone to drift to the background due to lack of discriminative target information, which evidences the effectiveness of target cues in videos for improving EgoVQL.

Due to limited space, we demonstrate more results, analysis, and ablation studies in the \emph{supplementary material}. 

\section{Conclusion}

In this paper, we present a novel approach, dubbed PRVQL, for improving EgoVQL via exploring crucial target knowledge from videos to refine features for robust localization. Our PRVQL is implemented as a multi-stage architecture. In each stage, two key modules, including AKG and SKG, are used to extract target appearance and spatial knowledge from the video. The knowledge from one stage is used as guidance to refine query and video features in the next stage, which are adopted for
learning more accurate knowledge for further feature refinement. Through this progressive process, PRVQL learns gradually improved knowledge, which in turn leads to better refined features for target localization in the final stage. To validate the effectiveness of PRVQL, we conduct experiments on Ego4D. Our experimental results show that PRVQL achieves state-of-the-art result and largely surpasses other methods, showing its efficacy.

{
\small
\bibliographystyle{ieeenat_fullname}
\bibliography{main}
}

\newpage
\appendix
\twocolumn
\section* {Supplementary Material}

For better understanding of this work, we provide additional details, analysis, and results as follow:

\begin{itemize}[label={}]
   \item \textbf{A. Detailed Architectures of Modules} \\
   In this section, we display the detailed architectures for the cross-attention block $\mathtt{CAB}$ and masked self-attention block $\mathtt{MaskedSA}$ in the main text.

   \vspace{0.5em}
    
   \item \textbf{B. Inference Details} \\
   We provide more details for the inference of PRVQL.

   \vspace{0.5em}

   \item \textbf{C. Additional Experimental Results} \\
   We offer more experimental results in this work, including more ablations and comparison of different method across different scales on the Ego4D dataset.

   \vspace{0.5em}

   \item \textbf{D. Visualization Analysis of  Spatial Knowledge} \\
   We provide visual analysis to show the learned target spatial knowledge.

   \vspace{0.5em}

   \item \textbf{E. More Qualitative Results} \\
   We demonstrate more qualitative results of our method
    for localizing the target object.
   
\end{itemize}

\section{Detailed Architectures of Modules}

\begin{figure}[!ht]
    \centering
    \includegraphics[width=0.88\linewidth]{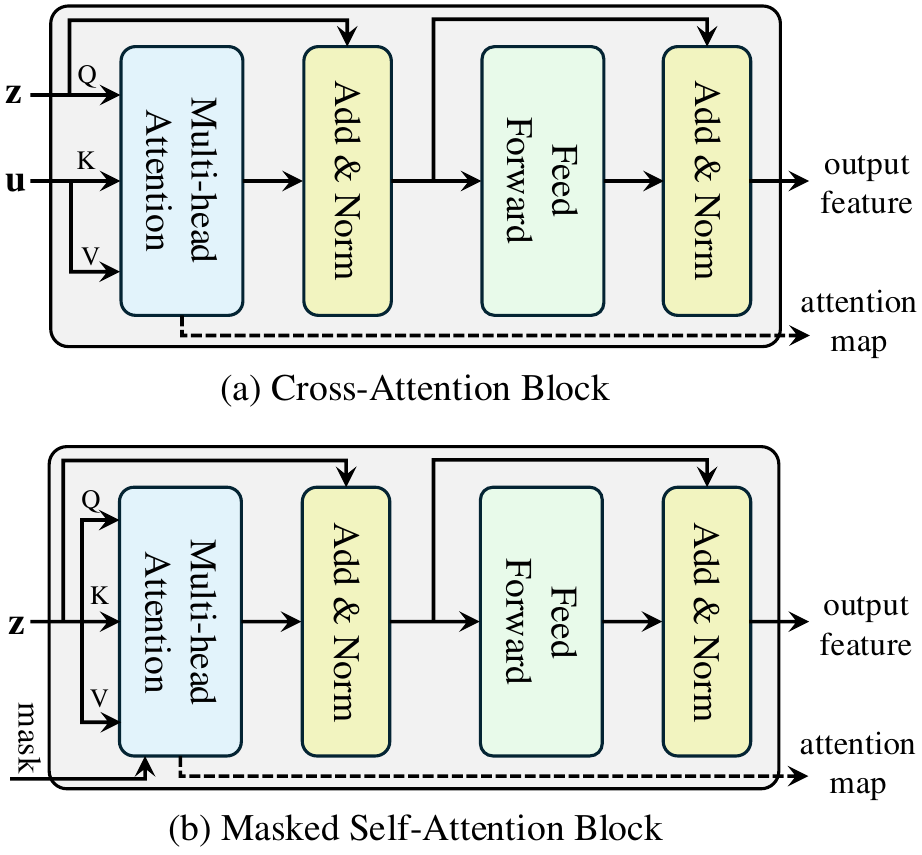}
    \vspace{-1mm}
    \caption{Detailed architectures of $\mathtt{CAB}$ and $\mathtt{MaskedSA}$.}
    \label{fig:support_fig3}\vspace{-4mm}
\end{figure}

In each stage of PRVQL, we adopt the cross-attention block $\mathtt{CAB}$ to fuse the query feature into the video feature and then utilize the masked self-attention block $\mathtt{MaskedSA}$ for further enhancing the video feature. The architectures of $\mathtt{CAB}$ and $\mathtt{MaskedSA}$ are shown in Fig.~\ref{fig:support_fig3}.

\section{Inference Details}

Similar to~\cite{jiang2024single}, for inference, we first predict the confidence scores for target occurrence in all frames. Given the scores, we then smooth them through a median filter with the kernel size of 1. After this, we perform peak detection on the smoothed scores. We detect the peak based on the highest score $h$ and use $0.79\cdot h$ as the threshold to filter non-confident peaks. Finally, we can determine a spatio-temporal tube that corresponds to the most recent peak as the prediction result. In order to detect start and end time of the tube, we threshold the confidences scores using the threshold of $0.585\cdot \tilde{h}$, where $\tilde{h}$ is the confidence score at the most recent peak.

\begin{table}[!t]
        \setlength{\tabcolsep}{7.5pt}
	\centering
    \caption{ Ablation studies on the parameter $\alpha$ in SKG.}\vspace{-2mm}
\renewcommand{\arraystretch}{1}
			\scalebox{0.92}{
				\begin{tabular}{ccccccc}
					\specialrule{1.5pt}{0pt}{0pt}
					\rowcolor{mygray} 
					& & tAP$_{25}$ & stAP$_{25}$ & rec\% &  Succ \\ \hline\hline
					\ding{182} & $\alpha$=0.4 & 0.33 & 0.26 & 47.27 & 57.24 \\
                    \ding{183} & $\alpha$=0.5 & \textbf{0.35} & \textbf{0.27} & \textbf{47.87} & \textbf{57.93} \\
                    \ding{184} & $\alpha$=0.6 & 0.31 & 0.26 & 46.34 & 55.97 \\
					\specialrule{1.5pt}{0pt}{0pt}
			\end{tabular}}
			\label{tab:merge_atten_lkg}
\end{table}

\begin{table}[!t]
        \setlength{\tabcolsep}{7.5pt}
	\centering
    \caption{Ablation studies on combination methods in QFR.}
    \vspace{-2mm}
			\renewcommand{\arraystretch}{1}
			\scalebox{0.89}{
				\begin{tabular}{ccccccc}
					\specialrule{1.5pt}{0pt}{0pt}
					\rowcolor{mygray} 
					& Method & tAP$_{25}$ & stAP$_{25}$ & rec\% &  Succ \\ \hline\hline
					\ding{182} & Addition & 0.31 & 0.23 & 46.57 & 56.37 \\
                    \ding{183} & Concatenation & 0.34 & 0.25 & 47.31 & 56.64 \\
                    \ding{184} & Cross-Attention & \textbf{0.35} & \textbf{0.27} & \textbf{47.87} & \textbf{57.93} \\
					\specialrule{1.5pt}{0pt}{0pt}
			\end{tabular}}
			\label{tab:merge_akg}
\end{table}

\begin{table}[!t]
    \setlength{\tabcolsep}{7.5pt}
    \centering
    \caption{Comparison on object of different scales in videos.}\vspace{-2mm}
    \scalebox{0.8}{
    \begin{tabular}{rccccc}
    \specialrule{1.5pt}{0pt}{0pt}
     \rowcolor{mygray} Method & Scale & tAP$_{25}$ & stAP$_{25}$ & rec$\%$ &Succ \\
     \cmidrule(r){1-1}\cmidrule(l){2-2}\cmidrule(l){3-6}
     CocoFormer        & \textit{small} & \textbf{0.067} & \textbf{0.030} & \textbf{19.565} & \textbf{21.113} \\
     VQLoC      & \textit{small} & 0.047          & 0.001          &   2.447          & 13.043\\
     PRVQL (ours)                       & \textit{small} & 0.036          & 0.004          &    2.351    & 16.087 \\
     \hline
     CocoFormer        & \textit{medium} & 0.206          & 0.127                  & 32.583  & 40.804 \\
     VQLoC     & \textit{medium} & 0.213          & 0.138                   & 33.738 &  44.719\\
     PRVQL (ours)                       & \textit{medium} & \textbf{0.261} & \textbf{0.179}  & \textbf{34.359} & \textbf{49.923}\\
     \hline
     CocoFormer        & \textit{large} & 0.338 & 0.271  & 40.737 & 56.164\\
     VQLoC      & \textit{large} & 0.454 & 0.387    & \textbf{53.635} &  67.680 \\
     PRVQL (ours)                       & \textit{large} & \textbf{0.469} & \textbf{0.396}  & 52.127 & \textbf{68.664} \\
     \specialrule{1.5pt}{0pt}{0pt}
    \end{tabular}}
    \label{tabel:scale}
\end{table}

\begin{figure}[!t]
    \centering
    \includegraphics[width=0.95\linewidth]{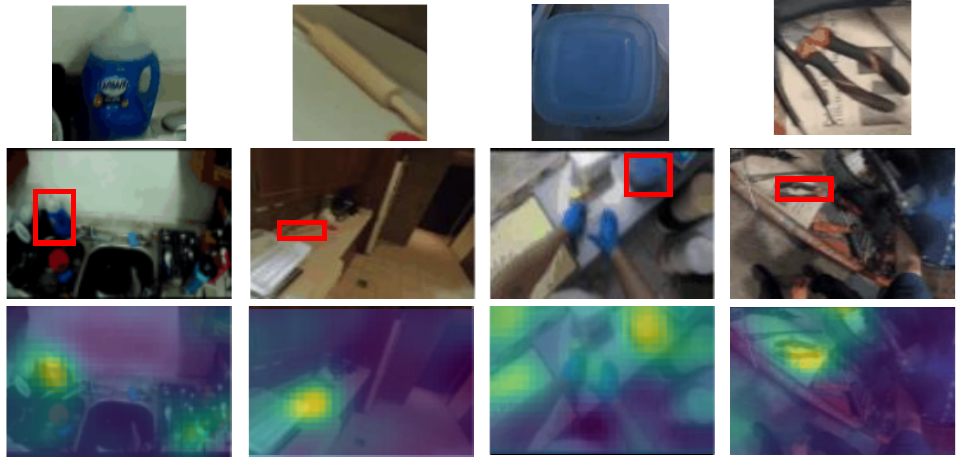}
    \vspace{-1mm}
    \caption{Visualization of target the spatial knowledge. First row: given visual query; second row: foreground target represented by red boxes; third row: learned target spatial knowledge.}
    \label{fig:support_fig2}\vspace{-3mm}
\end{figure}

\begin{figure}[!t]
    \centering
    \includegraphics[width=0.98\linewidth]{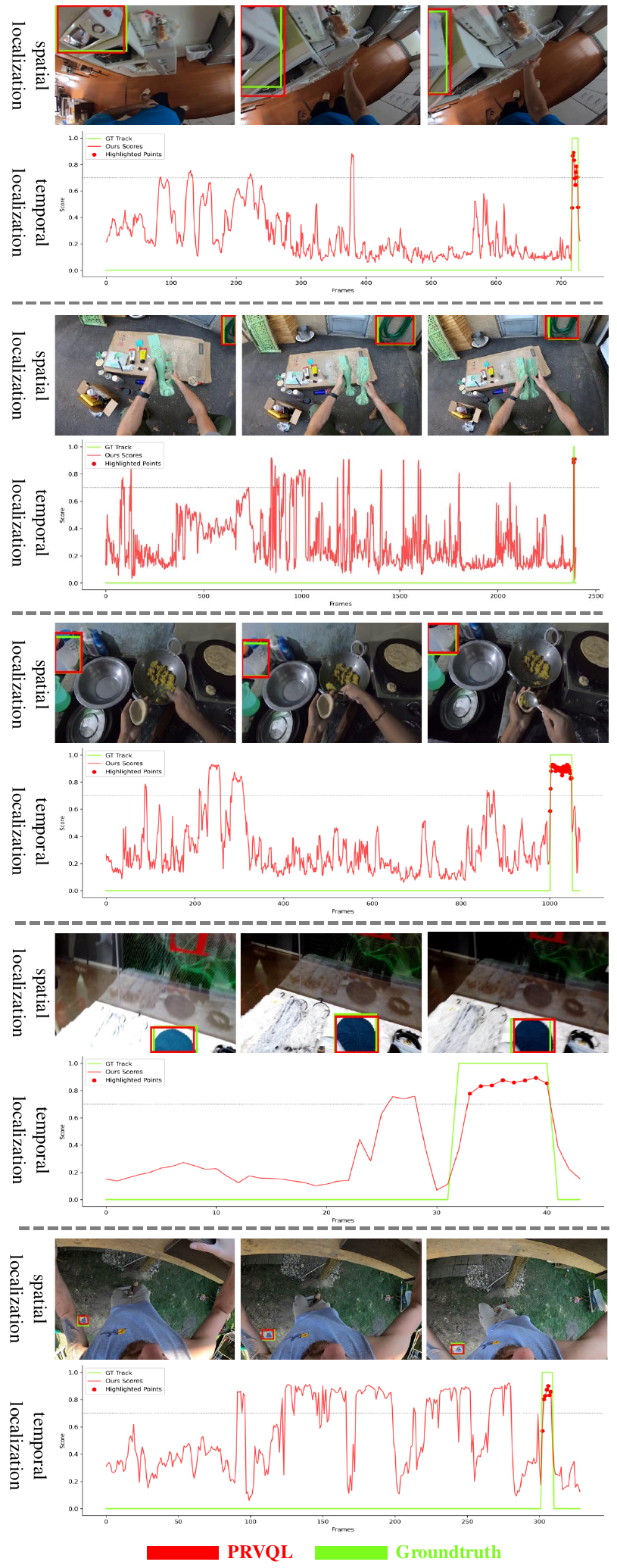}
    \vspace{-1mm}
    \caption{Qualitative results of our method.}
\label{fig:support_fig_qual}\vspace{-4mm}
\end{figure}

\section{Additional Experimental Results}

In this section, we show more ablation studies and comparison to other methods on the Ego4D validation set.

\vspace{0.3em}
\noindent
\textbf{Impact of Balance Parameter $\alpha$ in SKG.} The interpolated attention map $\varphi_{\text{int}}(\mathcal{T}_k^{\text{d}})$, obtained via bilinear interpolation from $\mathcal{T}_k^d$, is merged with $\mathcal{S}_k$ through a balance parameter $\alpha$. We conduct an ablation on $\alpha$ in Tab.~\ref{tab:merge_atten_lkg}. We can observe that, when setting $\alpha$ to 0.5, we show the best result (see \ding{183}). 

\vspace{0.3em}
\noindent
\textbf{Different Combination Methods in QFR.} In QFR, the appearance knowledge $\mathcal{K}^a_k$, obtained by AKG, is used to guide the refinement of query feature. In PRVQL, we use a cross-attention block to combine $\mathcal{K}^a_k$ and $\mathcal{Q}_k$ for achieving refinement. Besides cross-attention, we conduct experiments using other manners for refinement, including element-wise addition and concatenation, in Tab.~\ref{tab:merge_akg}. As shown in  Tab.~\ref{tab:merge_akg}, when using the cross-attention block for query feature refinement, we achieve the best performance (see \ding{184}).

 \vspace{0.3em}
\noindent
\textbf{Comparison in Different Scales.} Following~\cite{jiang2024single}, we provide comparison for objects of different scales in videos. As in ~\cite{jiang2024single}, the objects are categorized to three scales, including \emph{small} scale with target area in the range of [0, 64$^2$], \emph{medium} scale with the target area in the range (64$^2$, 192$^2$], and \emph{large} scale with target area greater than 192$^2$.  Tab.~\ref{tabel:scale} reports the comparison result. As in Tab.~\ref{tabel:scale}, we can observe that, CocoFormer performs better for small-scale objects. We argue that the reason is CocoFormer adopts higher-resolution images for localization and employs detector that is good at small object detection, while VQLoC and our method use downsampled frames for localization and do not specially deal with small objects. In comparison to CocoFormer and VQLoC, our PRVQL achieves better overall performance for medium- and large-scale objects, which shows the efficacy of target knowledge for robust target localization.

\section{Visualization Analysis of Spatial Knowledge}

The target spatial knowledge by SKG aims to explore target cues from videos for enhancing the target while suppressing background regions in video features. In PRVQL, we adopt the readily available attention maps to produce target spatial knowledge, which is shown in Fig.~\ref{fig:support_fig2}. From Fig.~\ref{fig:support_fig2}, we can see that, our spatial knowledge focuses more on the target object while less on the background, and thus can be applied to refine video features for better localization.

\section{More Qualitative Results}

In order to further validate the effectiveness of our PRVQL, we provide additional examples of target localization results in Fig.~\ref{fig:support_fig_qual}. From the shown visualizations, we can observe that, with the hlep of target knowledge, our method can accurately locate the target in both space and time.

\end{document}